\documentclass[11pt,a4paper]{article}
\usepackage[hyperref]{acl2020}
\usepackage{times}
\usepackage{latexsym}
\usepackage{enumitem}
\usepackage{multirow}
\usepackage{tikz}
\usepackage{algorithm}
\usepackage{algorithmic}

\renewcommand{\algorithmiccomment}[1]{\hfill$\triangleright$\textit{#1}}

\usepackage{tikz}
\usetikzlibrary{positioning,shadows,arrows,trees,shapes,fit}

\usepackage{amsmath}
\usepackage{amsfonts,bm}

\newtheorem{prop}{Proposition}

\newcommand{\sarrow}[1][4pt]{\!\mathrel{%
   \vcenter{\hbox{\rule[-.5\fontdimen8\textfont3]{#1}{\fontdimen8\textfont3}}}%
   \mkern-4mu\hbox{\usefont{U}{lasy}{m}{n}\symbol{41}}}\!}
\makeatletter   
\newcommand{\sveryshortarrow}[1][3pt]{\mathrel{%
    \vcenter{\hbox{\rule[-.5\fontdimen8\scriptfont3]
               {\scriptratio\dimexpr#1\relax}{\fontdimen8\scriptfont3}}}%
   \mkern-4mu\hbox{\let\f@size\sf@size\usefont{U}{lasy}{m}{n}\symbol{41}}}}
\makeatother

\def\eqref#1{equation~\ref{#1}}

\def\1{\bm{1}}

\def\ve{{\bm{e}}}

\def\vh{{\bm{h}}}

\def\vw{{\bm{w}}}

\def\m1{{\bm{1}}}

\def\mH{{\bm{H}}}

\def\mX{{\bm{X}}}

\DeclareMathAlphabet{\mathsfit}{\encodingdefault}{\sfdefault}{m}{sl}
\SetMathAlphabet{\mathsfit}{bold}{\encodingdefault}{\sfdefault}{bx}{n}

\def\gP{{\mathcal{P}}}
\def\gQ{{\mathcal{Q}}}

\def\gS{{\mathcal{S}}}
\def\gT{{\mathcal{T}}}
\def\gU{{\mathcal{U}}}

\newcommand{\Ls}{\mathcal{L}}

\newcommand{\ptr}{\rho}

\newcommand{\sptr}{sp}
\newcommand{\gptr}{gp}
\newcommand{\blb}{uc}
\newcommand{\glb}{gc}

\DeclareMathOperator*{\argmax}{arg\,max}

\DeclareMathOperator*{\pop}{pop}
\DeclareMathOperator*{\push}{push}

\usepackage{xspace}

\newcommand{\ie}{{\em i.e.,}\xspace}
\newcommand{\eg}{{\em e.g.,}\xspace}

\usepackage{url}
\usepackage{hyperref}
\usepackage{url}
\usepackage{graphicx}
\usepackage{amssymb}
\usepackage{comment}

\usepackage{caption}
\usepackage{subcaption}
\usepackage{booktabs}
\DeclareMathAlphabet{\pazocal}{OMS}{zplm}{m}{n}
\DeclareMathAlphabet{\pazocal}{OMS}{zplm}{m}{n}
\aclfinalcopy 
\def\aclpaperid{654} 

\title{Efficient Constituency Parsing by Pointing}

\author{Thanh-Tung Nguyen$^{\dagger}$$^\P$, Xuan-Phi Nguyen$^{\dagger}$$^\P$, Shafiq Joty$^\P$$^\S$, Xiaoli Li$^\dagger$\\
  $^\P$Nanyang Technological University \\
  $^\S$Salesforce Research Asia \\
  $^\dagger$Institute for Infocomm Research, A-STAR \\
  Singapore \\
  \texttt{\{ng0155ng@e.;nguyenxu002@e.;srjoty@\}ntu.edu.sg} 
  \\
  \texttt{xlli@i2r.a-star.edu.sg} 
}

\date{}

\begin{document}
\maketitle
\begin{abstract}

We propose a novel constituency parsing model that casts the parsing problem into a series of pointing tasks. Specifically, our model estimates the likelihood of a span being a legitimate tree constituent via the pointing score corresponding to the boundary words of the span. Our parsing model supports efficient top-down decoding and our learning objective is able to enforce structural consistency without resorting to the expensive CKY inference. The experiments on the standard English Penn Treebank parsing task show that our method achieves 92.78 F1 without using pre-trained models, which is higher than all the existing methods with similar time complexity. Using pre-trained BERT, our model achieves 95.48 F1, which is competitive with the state-of-the-art while being faster. Our approach also establishes new state-of-the-art in Basque and Swedish in the SPMRL shared tasks on multilingual constituency parsing.

\end{abstract}

\section{Introduction} \label{sec:intro}
Constituency or phrase structure parsing is a core task in natural language processing (NLP) with myriad downstream applications. Therefore, devising effective and efficient algorithms for parsing has been a key focus in NLP.

With the advancements in neural approaches, various neural  architectures have been proposed for constituency parsing as they are able to effectively encode the input tokens into dense vector representations while modeling the structural dependencies between tokens in a sentence. These include recurrent networks \citep{dyer-etal-2016-recurrent,stern-etal-2017-effective} and more recently self-attentive networks \citep{kitaev-klein-2018-constituency}.

The parsing methods can be broadly distinguished based on whether they employ a greedy transition-based algorithm or a globally optimized chart parsing algorithm. The transition-based parsers \citep{dyer-etal-2016-recurrent,cross-huang-2016-span, liu-zhang-2017-shift} generate trees autoregressively as a form of shift-reduce decisions. Though computationally attractive, the local decisions made at each step may propagate errors to subsequent steps which would suffer from exposure bias.

Chart parsing methods, on the other hand, learn scoring functions for subtrees and perform global search over all possible trees to find the most
probable tree for a sentence \citep{durrett-klein-2015-neural,gaddy-etal-2018-whats,kitaev-klein-2018-constituency,kitaev-etal-2019-multilingual}. In this way, these methods can ensure consistency in predicting structured output. The limitation, however, is that they run slowly at $\mathcal{O}(n^3)$ or higher time complexity.

\begin{figure}[t!]
\scalebox{0.9}{
\begin{tikzpicture}
    [font=\small, edge from parent, 
        edge from parent/.style={draw=blue!50,thick},
        level 1/.style={sibling distance=1.8cm},
        level 2/.style={sibling distance=1.8cm}, 
        level 3/.style={sibling distance=1.8cm},
        level 4/.style={sibling distance=1.8cm}, 
        level distance=1cm,
        ]
        \node (S) {S} 
            child { node [label=below:1] {She}
                }
            child { node {$\varnothing$}
                child { node {VP}
                    child { node [label=below:2] {enjoys }}
                    child { node {S-VP}
                        child { node [label=below:3] {playing}}
                        child  { node [label=below:4] {tennis}}
                        }
                    }
                child { node [label=below:5] {.}
                    }
                };
\node[align=left, above] at (-2.8,-5.1) {\textbf{Span Representation}};         
\node[align =left, above] at (-0.2,-5.5) {\textbf{$\gS(T) =$} \{((1, 5), S), ((2, 5), $\varnothing$), ((2, 4), VP), ((3, 4), S-VP)\}};                
\node[align=left, above] at (-2.7,-6.1) {\textbf{Pointing Representation}};         
\node[align =left, above] at (-0.1,-6.5) {\textbf{$\gP(T) =$} \{($1 \sarrow 5$,S), ($2\sarrow5$,$\varnothing$), ($3\sarrow4$,S-VP), ($4\sarrow2$,VP), ($5\sarrow1$,S)\}};
\end{tikzpicture}
}
\caption{A binarized constituency tree for the sentence ``She enjoys playing tennis.''. The node \emph{S-VP} is an example of a collapsed atomic label. We omit POS tags and singleton spans for simplicity. Below the tree, we show span and pointing representations of the tree.         
}
\label{fig:Tree2PointingFormat}
\end{figure}

In this paper, we propose a novel parsing approach that casts constituency parsing into a series of pointing problems (Figure \ref{fig:Tree2PointingFormat}). Specifically, our parsing model estimates the pointing score from one word to another in the input sentence, which represents the likelihood of the span covering those words being a legitimate phrase structure (\ie\ a subtree in the constituency tree). During training, the likelihoods of legitimate spans are maximized using the cross entropy loss. This enables our model to enforce structural consistency, while avoiding the use of structured loss that requires expensive $\mathcal{O}(n^3)$ CKY inference  \cite{gaddy-etal-2018-whats,kitaev-klein-2018-constituency}. The training in our model can be fully parallelized without requiring structured inference as in \cite{shen-etal-2018-straight,gomez-rodriguez-vilares-2018-constituent}. Our pointing mechanism also allows efficient top-down  decoding with a best and worse case running time of $\mathcal{O}(n \log n)$ and  $\mathcal{O}(n^2)$, respectively. 

In the experiments with English Penn Treebank parsing, our model without any pre-training achieves 92.78 F1, outperforming all existing methods with similar time complexity. With pre-trained BERT \citep{devlin-etal-2019-bert}, our model pushes the F1 score to 95.48, which is on par with the state-of-the-art \cite{kitaev-etal-2019-multilingual}, while supporting faster decoding. Our model also performs competitively on the multilingual parsing tasks in the SPMRL 2013/2014 shared tasks and establishes new state-of-the-art in Basque and Swedish. We will release our code at \href{https://ntunlpsg.github.io/project/parser/ptr-constituency-parser}{https://ntunlpsg.github.io/project/parser/ptr-constituency-parser}

\section{Model} \label{sec:model}

Similar to \citet{minimal-span-based-parsing}, we view constituency parsing as the problem of finding a set of labeled spans over the input sentence. Let $\gS(T)$ denote the set of \textit{labeled spans} for a parse tree $T$. Formally, $\gS(T)$ can be expressed as   
\begin{equation}
\gS(T) := \{ ((i_t, j_t), l_t) \}_{t=1}^{|\gS(T)|} \text{ for } i_t < j_t 
\end{equation}
\noindent where $|\gS(T)|$ is the  number of spans in the tree. Figure \ref{fig:Tree2PointingFormat}
shows an example constituency tree and its corresponding labeled span representation. 

Following the standard practice in parsing  \cite{gaddy-etal-2018-whats,shen-etal-2018-straight}, we convert the $n$-ary tree into a binary form and introduce a dummy label $\varnothing$ to spans that are not constituents in the original tree but created as a result of binarization. Similarly, the labels in unary chains corresponding to nested labeled spans  are collapsed into unique atomic labels, such as S-VP in Fig.  \ref{fig:Tree2PointingFormat}. 

Although our method shares the same ``span-based'' view with that of \citet{minimal-span-based-parsing}, our approach diverges significantly from their framework in the way we treat the whole parsing problem, and the representation and modeling of the spans, as we describe below.

\subsection{Parsing as Pointing}\label{subsec:parse_as_pointing}

In contrast to previous approaches, we cast parsing as a series of pointing decisions. For each index $i$ in the input sequence, the parsing model points it to another index $p_i$ in order to identify the tree span $(i, p_i)$, where $i \ne p_i$. Similar to Pointer Networks \cite{VinyalsNIPS2015}, each pointing mechanism is modeled as a multinomial distribution over the indices of the input tokens (or encoder states). However, unlike the original pointer network where a decoder state points to an encoder state, in our approach, every encoder state $\vh_i$ points to another encoder state $\vh_{p_i}$.

In this paper, we generally use $x \sarrow y$ to mean $x$ points to $y$. We will refer to the pointing operation either as a function of the encoder states (\eg\ $\vh_i \sarrow  \vh_{p_i}$) or simply the corresponding indices (\eg\ $i \sarrow p_i$). They both mean the same operation where the pointing function takes the encoder state $\vh_i$ as the query vector and points to $\vh_{p_i}$ by computing an attention distribution over all the encoder states.

Let $\gP(T)$ denote the set of pointing decisions derived from a tree $T$ by a transformation $\mathcal{H}$, \ie\ $\mathcal{H}: T \rightarrow \gP(T)$. For the parsing process to be valid, the transformation $\mathcal{H}$ and its inverse $\mathcal{H}'$ which transforms $\gP(T)$ back to $T$, should both have a one-to-one mapping property. Otherwise, the parsing model may confuse two different parse trees with the same pointing representation. In this paper, we propose a novel transformation that satisfies this property, as defined by the following proposition (proof provided in the Appendix).

\begin{prop}
\label{prop:1} 

Given a binary constituency tree $T$ for a sentence containing $n$ tokens, the transformation $\mathcal{H}$ converts it into a set of pointing decisions  $\gP(T) = \{(i\sarrow  p_i, l_i): i = 1, \ldots, n-1; i \ne p_i\}$ such that $(\min (i, p_i), \max (i, p_i))$ is the \textbf{largest} span that starts or ends at $i$, and $l_i$ is the label of the nonterminal associated with the span.
\end{prop}

To elaborate further, each pointing decision in $\gP(T)$ represents a specific span in $\gS(T)$. The pointing $i\sarrow  p_i$ is directional, while the span  that it represents $(i', j')$ is non-directional. In other words, there may exist position $i$ such that $i > p_i$, while $i' < j'\ \forall i', j' \in [1, n]$. {In fact, it is easy to see that if the token at index $i$ is a left-child of a subtree, the largest span involving $i$ starts at $i$, and in this case $i < p_i$ and $i'=i, j'=p_i$. On the other hand, if the token is a right-child of a subtree, the respective largest span ends at position $i$, in which case $i > p_i$ and $i'=p_i, j'=i$ (\eg\ see $4\sarrow2$ in Figure \ref{fig:Tree2PointingFormat}).} In addition, as the spans in $\gS(T)$ are unique, it can be shown that the pointing decisions in $\gP(T)$ are also distinct from one another (see Appendix for a proof by contradiction).

Given such pointing formulation, for every constituency tree, there exists a trivial case $(1\sarrow n, l_1)$ where $p_1=n$ and $l_1$ is generally `S'. Thus, to make our formulation more general with $n$ inputs and $n$ outputs and {convenient for the method description discussed later on,} we add another trivial case $(n\sarrow  1, l_1)$. With this generalization, we can represent the pointing decisions of any binary constituency tree $T$ as: 
\begin{equation}
\gP(T)=\{(i\sarrow  p_i, l_i): i= 1, \ldots, n; i \ne p_i \}    
\end{equation}
The pointing representation of the tree in Figure \ref{fig:Tree2PointingFormat} is given at the bottom of the figure. To illustrate, in the parse tree, the largest phrase that starts or ends at token 2 (`enjoys') is the subtree rooted at `$\varnothing$', which spans from 2 to 5. In this case, the span \textit{starts} at token 2. Similarly, the largest phrase that starts or ends at token 4 (`tennis') is the span ``enjoys playing tennis'', which is rooted at `VP'. In this case, the span \textit{ends} at token 4.

\begin{algorithm}[t!]
\small
  \caption{Convert binary tree to Pointing}
  \label{alg0}
  \begin{algorithmic}
  \REQUIRE Binary tree $T$ and its span representation $\gS(T)$
  \ENSURE Pointing representation $\gP(T)$
    \STATE $\gP(T)=[]$  \algorithmiccomment{Empty pointing list}
    \FOR{$\text{each leaf}_i$ in $T$}
        \STATE $node \leftarrow \text{leaf}_i$
        \STATE $(x, y) \leftarrow (i, i)$ \algorithmiccomment{Initialize current span, $x \leq y$}
        \STATE $l_i \leftarrow \varnothing$ \algorithmiccomment{Initialize label of current span}
        \WHILE {$x = i \text{ or } y = i$}  
            \STATE $p_i \leftarrow x + y - i$
            \STATE $l_i \leftarrow \text{node.label}$ \algorithmiccomment{The span's label}
            \STATE $\text{node} \leftarrow \text{node.parent}$
            \STATE $(x, y) \leftarrow \text{node.span}$ \algorithmiccomment{Span covered by node}
        \ENDWHILE   \algorithmiccomment{Until $i$ is no longer start/end point}
        \STATE $\push(\gP(T),(i\sarrow p_i,l_i))$
    \ENDFOR
    \RETURN $\gP(T)$
  \end{algorithmic}
\end{algorithm}

{Algorithm \ref{alg0} describes the procedure to convert a binary tree to its corresponding pointing representation. Specifically, from each leaf token $i$, the algorithm traverses upward along the hierarchy until the non-terminal node that does not start or end with $i$. In this way, the largest span starting or ending with $i$ can be identified.}

\subsection{Top-Down Tree Inference}\label{subsec:topdown_infer}

In the previous section, we described how to convert a constituency tree $T$ into a sequence of pointing decisions $\gP(T)$. We use this transformation to train the parsing model (described in detail in Sections \ref{subsec:model} - \ref{subsec:train}). During inference, given a sentence to parse, our decoder with the help of the parsing model predicts $\gP(T)$, from which we can construct the tree $T$. However, not all sets of pointings $\gP(T)$ guarantee the generation of a valid tree. For example, for a sentence with four (4) tokens, the pointing $\gP(T)=\{(1\sarrow  4, l_1), (2\sarrow  3, l_2), (3\sarrow  4, l_3), (4\sarrow  1, l_1)\}$ does not generate a valid tree because token `3' cannot belong to both spans $(2,3)$ and $(3,4)$. In other words, simply taking the $\argmax$ over the pointing distributions may not generate a valid tree. 

Our approach to decoding is inspired by the span-based approach of \citet{minimal-span-based-parsing}. In particular, to reduce the search space, we score for span identification (given by the pointing function) and label assignment separately.

\paragraph{Span Identification.} 
We adopt a top-down greedy approach formulated as follows. 
\begin{eqnarray}
    k^{*} &=  \argmax_{k} s_{\text{split}}(i, k, j) \label{eq:parsing}
\end{eqnarray}

\noindent where  $s_{\text{split}}(i, k, j)$ is the score of having a split-point at position $k$ ($i\le k < j$), as defined by the following equation.
\begin{eqnarray}
    s_{\text{split}}(i, k, j) = \ptr(k\sarrow i) + \ptr(k\!+\!1 \sarrow j) \label{eq:split}
\end{eqnarray}

\noindent where $\ptr(k\sarrow i)$ and $\ptr(k\!+\!1 \sarrow j)$ are the pointing scores (probabilities) for spans  $(i, k)$ and $(k\!+\!1, j)$, respectively. Note that the pointing scores are \emph{asymmetric}, meaning that $\ptr(i\sarrow j)$ may not be equal to $\ptr(j \sarrow i)$, because pointing from $i$ to $j$ is different from pointing from $j$ to $i$. This is different from previous approaches, where the score of a span is defined to be symmetric. We build a tree for the input sentence by computing Eq. \ref{eq:parsing} recursively starting from the full sentence span $(1,n)$.

\begin{figure*}[t!]
\centering
\begin{subfigure}[b]{0.49\linewidth}
\includegraphics[width=0.8\textwidth]{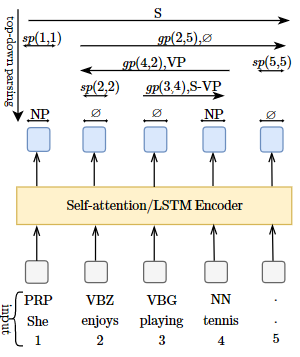}
\caption{Execution of pointing parsing algorithm}
\label{fig:topdownparsing}
\end{subfigure}
~
\begin{subfigure}[b]{0.4\linewidth}
\includegraphics[width=0.95\textwidth]{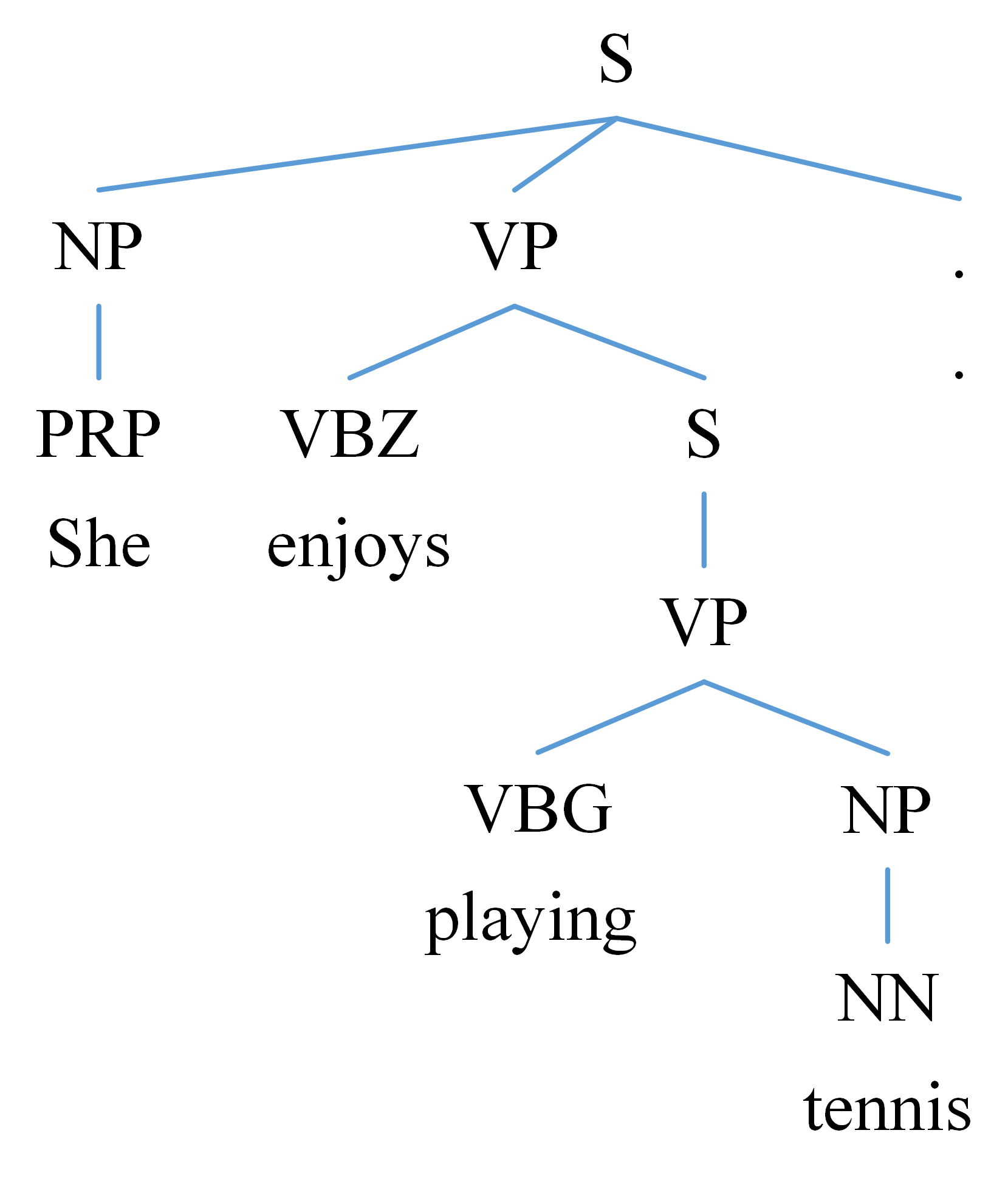}
\caption{Output parse tree.}
\label{fig_tree}
\end{subfigure}
\caption{Inferring the parse tree for a given sentence and its part-of-speech (POS) tags {(predicted by an external POS tagger)}. Starting with the full sentence span $(1,5)$ and its label $\text{S}$, we predict split point $1$ using the base ($\sptr$) and general ($\gptr$) pointing scores as per Eqn. \ref{eq:parsing}-\ref{eq:all}. The left singleton span $(1,1)$ is assigned with a label $\text{NP}$ and the right span $(2,5)$ is assigned with a label $\varnothing$ using the label classifier $\glb$ as per Eqn. \ref{eq:lab}. The recursion of splitting and labeling continues until the process reaches a terminal node. The label assignment for the unary spans is done by the $\blb$ classifier.} 

\label{fig:infer_model}
\end{figure*}

In the general case when $i\!<\!k\!<\!j-1$, our pointing-based parsing model should learn to assign high scores to the two spans $(i, k)$ and $(k\!+\!1, j)$, or equivalently the pointing decisions $k\sarrow i$ and $k\!+\!1\sarrow j$. 
However, the pointing formulation described so far omits the trivial \emph{self-pointing} decisions, which represent the \emph{singleton spans}. A singleton span is only created when the splitting decision splits an $n$-size span into a single-token span (singleton span) and a sub-span of size $n-1$, \ie\ when $k=i$ or $k=j\!-\!1$.
For instance, for the parsing process in Figure \ref{fig:topdownparsing}, the splitting decision at the root span $(1,5)$ results in a singleton span $(1,1)$ and a general span $(2,5)$. For this splitting decision, Eq. \ref{eq:parsing} requires the scores of $(1,1)$ and $(2,5)$. However, the set of pointing decisions $\gP(T)$ does not cover the pointing for $(1,1)$. This discrepancy can be resolved by modeling the singleton spans separately. To achieve that, we redefine Eq. \ref{eq:parsing} as follows:
\begin{equation}
\begin{aligned}[b]
    & \hspace{-0.5em}s_{\text{split}}(i, k, j) =\\
    & \hspace{-0.5em}\left\{
    	\begin{array}{ll}
 \hspace{-0.2em}\sptr(i \sarrow i) + \gptr(i \!+\! 1\sarrow j)   & \hspace{-0.6em} \mbox{if $k=i$ }\\ 
 \hspace{-0.2em}\gptr(j\!-\!1\sarrow i) + \sptr(j\sarrow j)   & \hspace{-0.6em}\mbox{if $k=j-1$ }\\    	    
 \hspace{-0.2em}\gptr(k\sarrow i) + \gptr(k \!+\!1\sarrow j)   & \hspace{-0.6em} \mbox{otherwise}           
    	\end{array}
    \right.
    \label{eq:all}
\end{aligned}
\end{equation}
where $\sptr$ and $\gptr$ respectively represent the scores for the singleton and general pointing functions (to be defined formally in Section \ref{subsec:model}). 

\textit{Remark on structural consistency.} It is important to note that since the pointing functions are defined to have a global structural property (\ie\ the largest span that starts/ends with $i$), our model inherently enforces structural consistency. The pointing formulation of the parsing problem also makes the training process simple and efficient; it allows us to train the model effectively with simple cross entropy loss (see Section \ref{subsec:train}).

\paragraph{Label Assignment.} {Label assignment} of spans is performed after every split decision. Specifically, as we split a span $(i,j)$ into two sub-spans $(i,k)$ and $(k\!+\!1,j)$ which corresponds to the pointing functions of $k\sarrow  i$ and $k\!+\!1\sarrow  j$, we perform the label assignments for the two new sub-spans as
\begin{equation}
  \begin{aligned}
  l_{k} =& \argmax_{l \in L} \glb (l|k)  \\
  l_{k+1} =& \argmax_{l \in L} \glb (l|k+1) \label{eq:lab}
  \end{aligned}
\end{equation}
where $\glb$ is the label classifier for any general (non-unary) span and $L$ is the set of possible non-terminal labels. 
Following \citet{shen-etal-2018-straight}, we use a separate {classifier $\blb$} for determining the labels of the unary spans, \eg\ the first layer of labels NP, $\varnothing$, $\ldots$, NP, $\varnothing$) in Figure \ref{fig:infer_model}. Also, note that the label assignment is done based on only the query vector (the encoder state that is used to point). 

Figure \ref{fig:infer_model} illustrates the top-down parsing process for our running example. It consists of a sequence of pointing decisions (Figure \ref{fig:topdownparsing}, top to bottom), which are then trivially converted to the parse tree (Figure \ref{fig_tree}). We also provide the pseudocode in Algorithm \ref{alg1}. Specifically, the algorithm finds the best split for the current span $(i, j)$ using the pointing scores and pushes the newly created sub-spans into the FIFO queue $\gQ$. The process terminates when there are no more spans to be split. Similar to  \citet{minimal-span-based-parsing}, our parsing algorithm has the worst and best case time complexities of $\mathcal{O}(n^2)$ and $\mathcal{O}(n\log n)$, respectively.

\begin{algorithm}[t!]
\small
  \caption{Pointing parsing algorithm}
  \label{alg1}
  \begin{algorithmic}
  \REQUIRE Sentence length $n$; pointing scores: $\gptr(i,j)$, $\sptr(i,j)$; label scores: $\glb(l|i)$, $\blb(l|i)$, $1\leq i\leq j \leq n, l \in L_g/L_u$
  \ENSURE Parse tree $\gT$
   \STATE $\gQ=[(1,n)]$  \algorithmiccomment{queue of spans}
   \STATE $\gS=[(1,n,\argmax_{l}{\glb}(l|1)]$ \algorithmiccomment{general spans, labels}
   \STATE $\gU= \hspace{-0.2em} \{ ((t, t), \argmax_{l}\blb(l|t))\}_{t=1}^{n}$  \algorithmiccomment{unary spans, labels}
   \WHILE {$\gQ \neq \varnothing$}
    \STATE $(i, j) = \pop(\gQ)$
    \IF{$j \leq i+1$}
        \STATE \textbf{Continue}
    \ENDIF
    \STATE $k^{*}=\argmax_{i \le k < j}s_\text{split}(i,k,j)$ 
    \hspace{-0.5em} \algorithmiccomment{using $\gptr, \sptr$}
    \IF{$k = i$}    
        \STATE $\push(\gQ,(i+1,j))$
        \STATE $\push(\gS,(i+1,j,\argmax_{l}\glb(l|i+1)))$
    \ELSIF{$k=j-1$}  
        \STATE $\push(\gQ,(i,j-1))$
        \STATE $\push(\gS,(i,j-1,\argmax_{l}\glb(l|j-1)))$
    \ELSE   
        \STATE $\push(\gQ,(i,k))$
        \STATE $\push(\gQ,(k+1,j))$
        \STATE $\push(\gS,(i,k,\argmax_{l}\glb(l|k)))$
        \STATE $\push(\gS,(k+1,j,\argmax_{l}\glb(l|k+1)))$
    \ENDIF
  \ENDWHILE
  \STATE $\gT=\gS \cup \gU$
  \end{algorithmic}
\end{algorithm}

\subsection{Model Architecture} \label{subsec:model}

We now describe the architecture of our parsing model: the sentence encoder, the pointing model and the labeling model.

\paragraph{Sentence Encoder.}
Given an input sequence of $n$ words $\mX= (x_1, \ldots, x_n)$, we first embed each word $x_i$ to its respective vector representation $e_i$ as:
\begin{equation}
    \ve_i = \ve_i^{\text{char}} + \ve_i^{\text{word}} + \ve_i^{\text{pos}}
\end{equation}
where $\ve_i^\text{char}$, $\ve_i^\text{word}$,  $\ve_i^\text{pos}$ are respectively the character, word, and part-of-speech (POS) embeddings of the word $x_i$. Following \citet{kitaev-klein-2018-constituency}, we use a character LSTM to compute the character embedding of a word. We experiment with both randomly initialized and pretrained word embeddings. If pretrained embeddings are used, the word embedding $\ve_i^{\text{word}}$ is the summation of the word's  randomly-initialized embedding and the pretrained embedding. The POS embeddings ($\ve_i^{\text{pos}}$) are randomly initialized.

The word representations ($\ve_i$) are then passed to a neural network based sequence encoder to obtain their hidden representations. Since our method does not require any specific encoder, one may use any encoder model, such as Bi-LSTM \citep{Hochreiter:1997} or self-attentive  encoder \citep{kitaev-klein-2018-constituency}. In this paper, unless otherwise specified, we use the self-attentive encoder model as our main sequence encoder {because of its efficiency with parallel computation}. The model is factorized into content and position information in both the  self-attention sub-layer and the feed-forward layer. Details about this factorization process is provided in \citet{kitaev-klein-2018-constituency}.

\paragraph{Pointing and Labeling Models.}
The results of the aforementioned sequence encoding process are used to compute the pointing and labeling scores. More formally,  the encoder network produces a sequence of $n$ latent vectors $\mH = (\vh_1, \ldots, \vh_n)$ for the input sequence $\mX= (x_1, \ldots, x_n)$. After that, we apply four (4) separate position-wise two-layer Feed-Forward Networks (FFN), formulated as $\text{FFN}(x) = \text{ReLU}(x W_1 + b_1)W_2 + b_2$, to transform $\mH$ into task-specific latent representations for the respective pointing and labeling tasks. 
\begin{align}
    \vh^{\gptr}_i=\text{FFN}_{\gptr}(\vh_i); \hspace{1em} \vh^{\sptr}_i=\text{FFN}_{\sptr}(\vh_i)\\
    \vh^{\glb}_i=\text{FFN}_{\glb}(\vh_i); \hspace{1em} \vh^{\blb}_i=\text{FFN}_{\blb}(\vh_i)
    \label{eqn:hidden_presentation}
\end{align}

Note that there is no parameter sharing between $\text{FFN}_{\gptr}$, $\text{FFN}_{\sptr}$, $\text{FFN}_{\glb}$ and $\text{FFN}_{\blb}$. The pointing functions are then modeled as the multinomial (or attention) distributions over the input indices for each input position $i$ as follows.
\begin{align}
    \gptr(i,k)=\frac{\exp(\vh^{\gptr}_i(\vh^{\gptr}_k)^T)}{\sum^{n}_{k=1} \exp(\vh^{\gptr}_i(\vh^{\gptr}_k)^T)}\\
    \sptr(i,k)=\frac{\exp(\vh^{\sptr}_i(\vh^{\sptr}_k)^T)}{\sum^{n}_{k=1} \exp(\vh^{\sptr}_i(\vh^{\sptr}_k)^T)}
    \label{eqn:softmax_pointing}
\end{align}

For label assignment functions, we simply feed the label representations $\mH^{\glb}=(\vh_1^{\glb},\ldots, \vh^{\glb}_n)$ and $\mH^{\blb}=(\vh_1^{\blb},\ldots, \vh^{\blb}_n)$ into the respective softmax classification layers as follows.
\begin{align}
    \glb(l|i) = \frac{\exp(\vh^{\glb}_i \vw^{\glb}_l)}{\sum^{|L_g|}_{l=1} \exp(\vh^{\glb}_i \vw^{\glb}_l)}\\
    \blb(l|i) = \frac{\exp(\vh^{\blb}_i \vw^{\blb}_l)}{\sum^{|L_u|}_{l=1} \exp(\vh^{\blb}_i \vw^{\blb}_l)}
    \label{eqn:softmax_unarylabel_spanlabel}
\end{align}
where $L_g$ and $L_u$ are the set of possible labels for the general and unary spans respectively, $\vw^{\glb}_l$ and $\vw^{\blb}_l$ are the class-specific trainable weight vectors.

\subsection{Training Objective} \label{subsec:train}

We train our parsing model by minimizing the total loss $\Ls_{total} (\theta)$ defined as:
\begin{eqnarray}
\Ls_{\text{total}} (\theta) = \hspace{-1em} &\Ls_{\gptr} (\theta_e, \theta_{\gptr}) +   \Ls_{\sptr} (\theta_e, \theta_{\sptr}) \nonumber \\ 
& + \Ls_{\glb} (\theta_e, \theta_{\glb}) + \Ls_{\blb} (\theta_e, \theta_{\blb})
\end{eqnarray}
\noindent where each individual loss is a cross entropy loss computed for the corresponding labeling or pointing task, and $\theta = \{\theta_e, \theta_{\gptr}, \theta_{\sptr}, \theta_{\glb}, \theta_{\blb}\}$ represents the overall model parameters; specifically, $\theta_e$ denotes the encoder parameters shared by all components, while $\theta_{\gptr}, \theta_{\sptr}, \theta_{\glb}$ and $\theta_{\blb}$ denote the separate parameters catering for the four pointing and labeling functions, $\gptr, \sptr, \glb$ and $\blb$, respectively.

\section{Experiments}\label{sec:experiment}
To show the effectiveness of our approach, we conduct experiments on English and Multilingual parsing tasks. For English, we use the standard Wall Street Journal (WSJ) part of the Penn Treebank (PTB) \cite{PTB:Marcus:1993}, whereas for multilingual, we experiment with seven (7) different languages from the SPMRL 2013-2014 shared task \citep{seddah-etal-2013-overview}: Basque, French, German, Hungarian, Korean, Polish and Swedish. 

For evaluation on PTB, we report the standard labeled precision (LP), labeled recall (LR), and labelled F1 computed by \texttt{evalb}\footnote{\url{http://nlp.cs.nyu.edu/evalb/}}. For the SPMRL datasets, we report labeled F1 and use the same setup in \texttt{evalb} as \citet{kitaev-klein-2018-constituency}. 

\subsection{English (PTB) Experiments}

\paragraph{Setup.} We follow the standard train/valid/test split, which uses sections 2-21 for training, section 22 for development and section 23 for evaluation. This gives 45K sentences for training, 1,700 sentences for development, and 2,416 sentences for testing. Following previous studies, our model uses POS tags predicted by the Stanford tagger \citep{Toutanova:2003:FPT:1073445.1073478}.


For our model, we adopt the self-attention encoder with similar hyperparameter details proposed by \citet{kitaev-klein-2018-constituency}. The character embeddings are of $64$ dimensions. For general and unary label classifiers ($gc$ and $uc$), the hidden dimension of the specific position-wise feed-forward networks is 250, while those for pointing functions ($gp$ and $sp$) have hidden dimensions of $1024$. Our model is trained using the Adam optimizer \citep{KingmaB14} with a batch size of $100$ sentences. Additionally, we use $100$ warm-up steps, within which we linearly increase the learning rate from $0$ to the base learning rate of $0.008$. Model selection for testing is performed based on the  labeled F1 score on the validation set.

\paragraph{Results for Single Models.} The experimental results on PTB for the models without pre-training are shown in Table \ref{tab:ptb_single}. As it can be seen, our model achieves an F1 of $92.78$, the highest among the models using \emph{top-down inference} strategies. Specifically, our method outperforms \citet{minimal-span-based-parsing} and \citet{shen-etal-2018-straight} by about $1.0$ point in F1-score. Notably, our model with LSTM encoder achieves an F1 of 92.26, which is still better than all the top-down parser methods.

\begin{table}[t]
\begin{center}
\resizebox{0.98\columnwidth}{!}{%
\setlength\tabcolsep{3.2pt}
\begin{tabular}{l|ccc} 
\toprule
{\bf Model}         & LR & LP & F1 \\
\midrule
\multicolumn{4}{c}{\textbf{Top-Down Inference}}    \\
\citet{minimal-span-based-parsing} & 93.20 & 90.30 & 91.80 \\
\citet{shen-etal-2018-straight}  & 92.00 & 91.70 & 91.80 \\
{Our Model}  & \textbf{92.81} & \textbf{92.75} & \textbf{92.78} \\
\midrule
\multicolumn{4}{c}{\textbf{CKY/Chart Inference} }  \\
\citet{gaddy-etal-2018-whats}  & - & - & 92.10 \\
\citet{kitaev-klein-2018-constituency} &93.20 &93.90&93.55\\
\midrule
\multicolumn{4}{c}{\textbf{Other Approaches}}   \\
\citet{gomez-rodriguez-vilares-2018-constituent} &- &- &90.7\\
\citet{liu-zhang-2017-shift} &- &- &91.8 \\
\citet{stern-etal-2017-effective} &92.57&92.56&92.56\\
\citet{ZhouZ19} &93.64 &93.92& 93.78\\
\bottomrule
\end{tabular}
\setlength\tabcolsep{6pt}
}
\caption{\label{tab:ptb_single}Results for single models (no pre-training) on the PTB WSJ test set, Section 23.}
\label{table:ptb_single_model}
\end{center}
\end{table}

On the other hand, while \citet{kitaev-klein-2018-constituency} and \citet{ZhouZ19} achieve higher F1 score, their inference speed is significantly slower than ours because of the use of CKY based algorithms, which run at $\mathcal{O}(n^3)$ time complexity for \citet{kitaev-klein-2018-constituency} and $\mathcal{O}(n^5)$ for \citet{ZhouZ19}. Furthermore, their training objectives involve the use of structural hinge loss, which requires online CKY inference during training. This makes their training time considerably slower than that of our method, which is trained directly with span-wise cross entropy loss.

In addition, \citet{ZhouZ19} uses external supervision (\emph{head} information) from the dependency parsing task. Dependency parsing models, in fact, have a strong resemblance to the pointing mechanism that our model employs \citep{ma-etal-2018-stack}. As such, integrating dependency parsing information into our model may also be beneficial. We leave this for future work.

\paragraph{Results with Pre-training} Similar to \citet{kitaev-klein-2018-constituency} and \citet{kitaev-etal-2019-multilingual}, we also evaluate our models with BERT \cite{devlin-etal-2019-bert} embeddings
. Following them in the inclusion of contextualized token representations, we adjust the number of self-attentive layers to $2$ and the base learning rate to $0.00005$.

As shown in Table \ref{table:ptb_pretrained_model}, our model achieves an F1 score of 95.48, which is on par with the state-of-the-art models. However, the advantage of our method is that it is faster than those methods. Specifically, our model runs at $\mathcal{O}(n^2)$ worst-case time complexity, while that of \citet{kitaev-etal-2019-multilingual} is $\mathcal{O}(n^3)$. Comparison on parsing speed is discussed in the following section.

\begin{table}[t]
\begin{center}
\resizebox{0.98\columnwidth}{!}{%
\setlength\tabcolsep{1.2pt}
\begin{tabular}{lc} 
\toprule
{\bf Model}         & {\bf F1} \\
\midrule
Our model $\text{\small{BERT}}_\text{BASE-uncased}$& 95.34 \\
Our model $\text{\small{BERT}}_\text{LARGE-cased}$& 95.48 \\
\midrule
\citet{kitaev-klein-2018-constituency} $\text{\small{ELMO}}$ & 95.13\\
\citet{kitaev-etal-2019-multilingual} $\text{\small{BERT}}_\text{LARGE-cased}$ &95.59\\
\bottomrule
\end{tabular}
\setlength\tabcolsep{6pt}
}
\caption{\label{tab:ptb_pretrain}Restuls on PTB WSJ test set with pretraining.}
\label{table:ptb_pretrained_model}
\end{center}
\end{table}

\paragraph{Parsing Speed Comparison.}

In addition to parsing performance in F1 scores, we also compare our parser against the previous neural approaches in terms of parsing speed. We record the parsing timing over {2416} sentences of the PTB test set with batch size of {1}, on a machine with NVIDIA GeForce GTX 1080Ti GPU and Intel(R) Xeon(R) Gold 6152 CPU. This setup is comparable to the setup of \citet{shen-etal-2018-straight}.

\begin{table}[t]
\begin{center}
\resizebox{0.7\columnwidth}{!}{%
\setlength\tabcolsep{1.2pt}
\begin{tabular}{lc} 
\toprule
{\bf Model}         & \# sents/sec \\

\midrule
\citet{petrov-klein-2007-improved}      &6.2 \\
\citet{zhu-etal-2013-fast}        &89.5 \\
\citet{liu-zhang-2017-shift}  &79.2 \\
\citet{minimal-span-based-parsing}            &75.5 \\
\citet{kitaev-klein-2018-constituency} & 94.40\\
\citet{shen-etal-2018-straight}             &111.1 \\

\midrule

Our model       &130.2   \\

\bottomrule
\end{tabular}
\setlength\tabcolsep{6pt}
}
\caption{\label{table_speed} Parsing speed for different models computed on the PTB WSJ test set.}
\label{table:table_speed}
\end{center}
\end{table}

\begin{table*}[t]
\begin{center}
\resizebox{1.9\columnwidth}{!}{%
\begin{tabular}{l|cccccccc} 
\toprule
{\bf Model}         & {\bf Basque} & {\bf French} & {\bf German}  & {\bf  Hebrew}  & {\bf Hungarian} & {\bf Korean} & {\bf Polish} & {\bf Swedish}\\
\midrule
\cite{spmrl2014}  &88.24 &82.53 &81.66 & 89.80&91.72 &83.81 &90.50 &85.50\\
\cite{coavoux-crabbe-2017-multilingual} & 88.81 & 82.49 & 85.34 & 89.87 & 92.34 & 86.04 & 93.64 & 84.0 \\
\cite{kitaev-klein-2018-constituency}   & 89.71 & 84.06 & 87.69 & 90.35 & 92.69 & 86.59 & 93.69 & 84.45 \\
Our Model & 90.23 & 82.20 & 84.91 & 90.63 & 91.07 & 85.36 & 93.99 & 86.87  \\
\bottomrule
\end{tabular}
\setlength\tabcolsep{6pt}
}
\caption{\label{tab:spmrl_dataset}SPMRL experiment single model test.}
\label{table:spmrl_single_model}
\end{center}
\end{table*}

\begin{table*}[t!]
\begin{center}
\resizebox{1.9\columnwidth}{!}{%
\begin{tabular}{l|ccccccccc} 
\toprule
{\bf Model}         
& {\bf Basque} & {\bf French} & {\bf German} & {\bf  Hebrew} & {\bf Hungarian} & {\bf Korean} & {\bf Polish} & {\bf Swedish}\\
\midrule
\cite{kitaev-etal-2019-multilingual}  & 91.63 & 87.43 & 90.20 &92.99 & 94.90 & 88.80 &  96.36 & 88.86 \\
Our model & 92.02 & 86.69 & 90.28 &93.67& 94.24 & 88.71 & 96.14 & 89.10  \\

\bottomrule
\end{tabular}
\setlength\tabcolsep{6pt}
}
\caption{\label{tab:spmrl_dataset_bert}SPMRL experiment pre-trained model test (with pretraining).}
\label{table:spmrl_bert_model}
\end{center}
\end{table*}

\begin{table}[t!]
\begin{center}
\resizebox{0.8\columnwidth}{!}{%
\begin{tabular}{l|ccc} 
\toprule
{\bf Language} & Train & Valid & Test \\
\midrule
{\bf Basque} & 7,577 &   948 & 946 \\
{\bf French} & 14,759 &  1,235 &  2,541 \\
{\bf German} & 40,472 & 5,000  & 5,000  \\
{\bf Hebrew} & 5,000 & 500  & 716  \\
{\bf Hungarian} & 8,146 & 1,051 & 1,009\\
{\bf Korean} & 23,010 &  2,066  & 2,287\\ 
{\bf Polish} & 6,578  &  821 & 822 \\
{\bf Swedish} & 5,000 & 494 & 666 \\
\bottomrule
\end{tabular}
}
\caption{\label{tab:spmrl_dataset_stat}SPMRL Multilingual dataset split.}
\label{table:spmrl_corpus_stats}
\end{center}
\end{table}

As shown in Table \ref{table_speed}, our parser outperforms \citet{shen-etal-2018-straight} by 19 more sentences per second, despite the fact that our parsing algorithm runs at $\mathcal{O}(n^2)$ worse-case time complexity while the one used by \citet{shen-etal-2018-straight} can theoretically run at {$\mathcal{O}(n\log n)$ time complexity}. To elaborate further, the algorithm presented in \citet{shen-etal-2018-straight} can only run at $\mathcal{O}(n^2)$ complexity. To achieve $\mathcal{O}(n\log n)$ complexity, it needs to sort the list of syntactic distances, which the provided code\footnote{\url{https://github.com/hantek/distance-parser}} does not implement.
In addition, the speed up for our method can be attributed to the fact that our algorithm (see Algorithm \ref{alg1}) uses a \emph{while loop}, while the algorithm of \citet{shen-etal-2018-straight} {has} many recursive function calls. Recursive algorithms tend to be less empirically efficient than their equivalent while/for loops in handling low-level memory allocations and function call stacks.

\subsection{SPMRL Multilingual Experiments}

\paragraph{Setup.} Similar to the English PTB experiments, we use the predicted POS tags from external taggers (provided in the SPMRL datasets). The train/valid/test split is reported in Table \ref{table:spmrl_corpus_stats}. For single model evaluation, we use the identical hyper-parameters and optimizer setups as in English PTB. For experiments with pre-trained models, we use the multilingual BERT \citep{devlin-etal-2019-bert}, which was trained jointly on 104 languages.


\paragraph{Results.} The results for the single models are reported in Table \ref{table:spmrl_single_model}. We see that our model achieves the highest F1 score in Basque and Swedish, which are higher than the baselines by $0.52$ and $1.37$ respective in F1. Our method also performs competitively with the previous state-of-the-art methods on other languages.

Table \ref{table:spmrl_bert_model} reports the performance of the models using pre-trained BERT. Evidently, our method achieves state-of-the-art results in Basque and Swedish, and performs on par with the previous best method by \citet{kitaev-etal-2019-multilingual} in the other five languages. Again, note that our method is considerably faster and easier to train than the method of \citet{kitaev-etal-2019-multilingual}. 

\section{Related Work} \label{sec:rel}
Prior to the neural tsunami in NLP, parsing methods typically model correlations in the \emph{output space} through probabilistic context-free grammars (PCFGs) on top of sparse (and discrete) input representations either in a generative regime \cite{klein-manning-2003-accurate} or a discriminative regime \citep{finkel-etal-2008-efficient} or a combination of both \cite{charniak-johnson-2005-coarse}. Beside the chart parser approach, there is also a long tradition of transition-based parsers \cite{sagae-lavie-2005-classifier}

Recently,  however, with the advent of powerful neural encoders such as LSTMs \cite{Hochreiter:1997}, the focus has been switched more towards effective modeling of correlations in the \emph{input's latent space,} as the output structures are nothing but a function of the input \citep{gaddy-etal-2018-whats}. Various neural network models have been proposed to effectively encode the dense input representations and correlations, and have achieved  state-of-the-art parsing results. To enforce the structural consistency, existing neural parsing methods either employ a transition-based algorithm \citep{dyer-etal-2016-recurrent,liu-zhang-2017-shift, tetra-tagging-2019} or a globally optimized chart-parsing algorithm \citep{gaddy-etal-2018-whats,kitaev-klein-2018-constituency}.

Meanwhile, researchers also attempt to convert the constituency parsing problem into tasks that can be solved in alternative ways. For instance, \citet{fernandez-gonzalez-martins-2015-parsing} transform the phrase structure into a special form of dependency structure. Such a dependency structure, however, requires certain corrections while converting back to the corresponding constituency tree. \citet{gomez-rodriguez-vilares-2018-constituent} and \citet{shen-etal-2018-straight} propose to map the constituency tree for a sentence of $n$ tokens into a sequence of $n-1$ labels or scalars based on the depth or height of the lowest common ancestors between pairs of consecutive tokens. In addition, methods like \cite{NIPS2015_5635,NIPS2017_7181} apply the sequence-to-sequence framework to ``translate'' a sentence into the linearized form of its constituency tree. While being trivial and simple, parsers of this type do not guarantee structural correctness, because the syntax of the linearized form is not constrained during tree decoding.

Our approach differs from previous work in that it represents the constituency structure as a series of pointing representations and has a relatively simpler cross entropy based learning objective. The pointing representations can be computed in parallel, and can be efficiently converted into a full constituency tree using a top-down algorithm. 
Our pointing mechanism shares certain similarities with the Pointer Network \citep{VinyalsNIPS2015}, but is distinct from it in that our method points a word to another word within the same encoded sequence.

\section{Conclusion}
We have presented a novel constituency parsing method that is based on a pointing mechanism. Our method utilizes an efficient top-down decoding algorithm that uses pointing functions for scoring possible spans. The pointing formulation inherently captures global structural properties and allows efficient training with cross entropy loss. With experiments we have shown that our method  outperforms all existing top-down methods on the English Penn Treebank parsing task. Our method with pre-training rivals the state-of-the-art method, while being faster than it. On multilingual constituency parsing, it also establishes new state-of-the-art in Basque and Swedish.

\section*{Acknowledgments}

We would like to express our gratitude to the anonymous reviewers for their insightful feedback on our paper. Shafiq Joty would like to
thank the funding support from his Start-up Grant
(M4082038.020).

\bibliographystyle{acl_natbib}
\bibliography{refs.bib}
 \section*{Appendix}
    
\paragraph{Proof of Proposition 1} Given $\gP(T) = \{(i\sarrow  p_i, l_i): i = 1, \ldots, n-1; i \ne p_i\}$, generated from tree $T$ (here we omit the unary leaves and POS-tags), we at first define the inverse $\mathcal{H}'$ as follows:
\begin{equation}
\begin{aligned}[b]
\mathcal{H}'(\gP(T))&=\{((min(i,p_i),max(i,p_i)),l_i): \\
&i = 1, \ldots, n-1\} \nonumber
\end{aligned}
\end{equation}
We would prove $\mathcal{H}'(\gP(T))=T$\\
A binary tree $T$ has exactly $n$$-$$1$ internal nodes (or spans). 
It is noteworthy to mention that for each pointing $(i \sarrow p_i,l_i)$,  $((\min (i, p_i), \max (i, p_i)),l_i)$ is a span in $T$. As we consider $i$ from $1$ to $n-1$, there are totally at most n-1 such spans in $\mathcal{H}'(\gP(T))$(we do not know whether these spans are not be distinct). Therefore, if we can prove that all $((\min (i, p_i), \max (i, p_i)),l_i)$ spans are distinct for $i= 1, \ldots, n-1$, $\mathcal{H}'(\gP(T))$ will cover all the span in $T$, therefore, $\mathcal{H}'(\gP(T))=T$. We prove this by contradiction.\\
Assume that there exist $i, j \in \{1, \ldots, n-1 \}$ such that $(\min(i,p_i), \max(i,p_i)) =(\min(j,p_j), \max(j,p_j))$ for $j\neq i$. First, if $p_i=n$, then according to the above condition, $(\min(j,p_j), \max(j,p_j))= (\min(i,n), \max(i,n))=(i,n)$. This means, either $j=n$ or $j=i$, which contradicts with our initial assumption that $j\neq i$ and $j \in \{1, \ldots, n-1 \}$. So, $p_i$ cannot be equal to $n$. Similarly, we can prove that $p_j$ also cannot be equal to $n$. Thus, we can conclude that $p_i,p_j \in \{1, \ldots, n-1 \}$.
Now, without loss of generality, let us assume that $j > i$. With this assumption, the two spans will be identical if and only if $p_i = j$ and $p_j = i$. In this case, the span $(i,j)$ would be the largest span that starts with $i$ and ends at $j$. However, since $1 \leq i < j \leq n-1$, the span $(i,j)$ must be a left or right child of another (parent) span. If $(i,j)$ is the left child, then the parent span needs to start with $i$, making it larger than $(i,j)$. This contradicts to the property that $(i,j)=(i,p_i)$ is the largest span that starts or ends at $i$. Similarly, if $(i,j)$ is the right child, then the parent span needs to end at $j$, making it larger than $(i,j)$. This again contradicts to the property that $(j,i)=(j,p_j)$ is the largest span that starts or ends at $j$. \\
In conclusion, we have $\mathcal{H}'(\gP(T))=T$. This would guarantee that $\mathcal{H}$ and $\mathcal{H}'$ are one-to-one: If there exist $T_1,T_2$ such that $\gP(T_1)=\gP(T_2)$, we would have $\mathcal{H}'(\gP(T_1))=\mathcal{H}'(\gP(T_2))$ or $T_1=T_2$.If there exist $T_1,T_2$ such that $\mathcal{H}'(\gP(T_1))=\mathcal{H}'(\gP(T_2))$, we would have $T_1=T_2$.
\end{document}